\documentclass[lettersize,journal]{IEEEtran}

\usepackage[utf8]{inputenc}
\usepackage[T1]{fontenc}

\usepackage{amsmath}
\usepackage{amsfonts}
\usepackage{bm}
\usepackage{mathtools}
\usepackage{nicefrac}

\usepackage{graphicx}
\usepackage{subfig} 

\usepackage{array}
\usepackage{booktabs}
\usepackage{multirow}
\usepackage{colortbl}
\usepackage{makecell}

\usepackage{algorithm}
\usepackage{algorithmic}

\usepackage{xcolor}

\definecolor{lightblue}{RGB}{173,216,230}
\newcommand{\lightbluecell}[1]{\cellcolor{lightblue}{#1}}

\definecolor{palepink}{RGB}{255,224,230}
\newcommand{\palepinkcell}[1]{\cellcolor{palepink}{#1}}

\definecolor{orange}{RGB}{255,165,0}
\newcommand{\orangecell}[1]{\cellcolor{orange}{#1}}

\usepackage[nocompress]{cite}
\usepackage{hyperref}
\usepackage[capitalise]{cleveref}

\hypersetup{
	colorlinks = true,
	citecolor  = blue,
	linkcolor  = blue,
	urlcolor   = black,
	filecolor  = magenta
}

\usepackage{microtype}
\usepackage{textcomp}
\usepackage{comment}
\usepackage{verbatim}
\usepackage{relsize}

\usepackage{orcidlink}
\usepackage{bbding}
\usepackage{wasysym}

\usepackage{stfloats}
\usepackage{balance}

\def\BibTeX{{\rm B\kern-.05em{\sc i\kern-.025em b}%
\kern-.08em T\kern-.1667em\lower.7ex\hbox{E}\kern-.125emX}}

\begin{document}

\title{Enhancing Visual Question Answering with Multimodal LLMs via Chain-of-Question Guided Retrieval-Augmented Generation}

\author{
Quanxing~Xu\orcidlink{0009-0008-4354-8371},
Ling~Zhou\orcidlink{0000-0002-8313-5749},
Xian~Zhong\orcidlink{0000-0002-5242-0467},~\IEEEmembership{Senior~Member,~IEEE},
Xiaohua~Huang\orcidlink{0000-0001-8897-3517},~\IEEEmembership{Senior~Member,~IEEE},
Rubing~Huang\orcidlink{0000-0002-1769-6126},~\IEEEmembership{Senior~Member,~IEEE},
and~Chia-Wen~Lin\orcidlink{0000-0002-9097-2318},~\IEEEmembership{Fellow,~IEEE}
 
\thanks{Manuscript received February 25, 2026. This work was supported in part by the Science and Technology Development Fund of Macau, Macao SAR, under Grants 0035/2023/ITP1 and 0021/2023/RIA1, the National Natural Science Foundation of China under Grant 62271361, and the Hubei Provincial Key Research and Development Program under Grant 2024BAB039. (\emph{Corresponding authors: Ling Zhou and Xian Zhong}.)}

\thanks{Quanxing Xu and Ling Zhou are with the School of Computer Science and Engineering, Macau University of Science and Technology, Macao SAR 999078, China (e-mail: 3230002299@student.must.edu.mo; lzhou@must.edu.mo).}

\thanks{Xian Zhong is with the Hubei Key Laboratory of Transportation Internet of Things, School of Computer Science and Artificial Intelligence, Wuhan University of Technology, Wuhan, Hubei 430070, China, and also with the State Key Laboratory of Maritime Technology and Safety, Wuhan University of Technology, Wuhan 430063, China (e-mail: zhongx@whut.edu.cn).}


\thanks{Xiaohua Huang is with the Oulu School, Nanjing Institute of Technology, Nanjing 210096, China (e-mail: xiaohuahwang@gmail.com).}

\thanks{Rubing Huang is with the School of Computer Science and Engineering, Macau University of Science and Technology, Macao SAR 999078, China, and also with the Zhuhai MUST Science and Technology Research Institute, Macau University of Science and Technology, Zhuhai, Guangdong 519099, China (e-mail: rbhuang@must.edu.mo).}

\thanks{Chia-Wen Lin is with the Department of Electrical Engineering, National Tsing Hua University, Hsinchu 30013, Taiwan (e-mail: cwlin@ee.nthu.edu.tw).}
}

\markboth{IEEE TRANSACTIONS ON IMAGE PROCESSING, 2026}%
{How to Use the IEEEtran \LaTeX Templates}

\maketitle

\begin{abstract}
With advances in multimodal research and deep learning, Multimodal Large Language Models (MLLMs) have emerged as a powerful paradigm for a wide range of multimodal tasks. As a core problem in vision-language research, Visual Question Answering (VQA) has increasingly employed MLLMs to improve performance, particularly in open-domain settings where external knowledge is essential. In this work, we aim to further enhance retrieval-based VQA by more effectively integrating MLLMs with structured reasoning and knowledge acquisition. We introduce a logical prompting strategy that fuses Chain-of-Thought (CoT) reasoning with Visual Question Decomposition (VQD), termed CoVQD, to guide retrieval toward more accurate and relevant knowledge for MLLM inference. Building on this idea, we propose a new framework, \underline{\textbf{C}}oVQD-\underline{\textbf{g}}uided \underline{\textbf{RAG}} (CgRAG), which enables MLLMs to access more comprehensive and coherent external knowledge while benefiting from structured visual-text reasoning guidance, thereby improving generalization and reliability in complex cross-domain VQA scenarios. Extensive experiments on \textsc{E-VQA}, \textsc{InfoSeek}, and \textsc{OKVQA} benchmarks demonstrate the effectiveness of the proposed method. 

\end{abstract}


\maketitle

\section{Introduction}
\label{sec:intro}

\IEEEPARstart{V}{ision}--language tasks serve as representative benchmarks for evaluating models’ capabilities in multimodal learning and visual-linguistic understanding, including visual storytelling~\cite{zhong1}, video captioning~\cite{zhong2,zhong3}, and Visual Question Answering (VQA)~\cite{antol2015}. As one of the most fundamental vision-language tasks, VQA requires generating accurate natural language answers given an image and a question, and has therefore attracted substantial attention in recent years. This growing interest reflects a broader shift in the image processing community from conventional ``bucketed’’ recognition problems toward more complex multimodal reasoning challenges~\cite{antol2015,goyal2017}.

Knowledge-based VQA (KBVQA)~\cite{kvqa}, an important subtask of VQA, explicitly requires information beyond an image's visual content. Early KBVQA approaches were predominantly retrieval-based, leveraging structured knowledge bases such as Wikipedia or ConceptNet. With the advent of large language models (LLMs), more recent methods increasingly rely on frozen LLMs as implicit repositories of world knowledge. Alongside advances in multimodal research, Multimodal LLMs (MLLMs) have further expanded this paradigm, ushering in a new era of multimodal generation and reasoning. Nevertheless, how to more fully unlock the reasoning potential of large models, improve domain adaptability, and mitigate predictive hallucinations remains a pressing challenge in open-domain VQA.

\begin{figure}[!t]
	\centering
	\includegraphics[width = \linewidth]{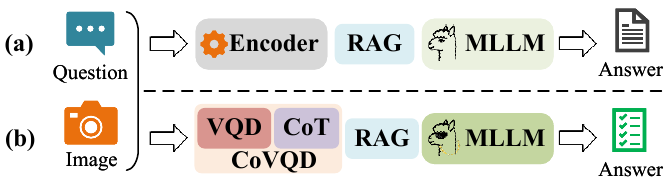}
	\caption{\textbf{Comparison between prior retrieval-based VQA with MLLMs and the proposed CgRAG framework.} With an MLLM fine-tuned by liDPO, VQD and CoT are fused to guide fine-grained RAG, yielding an enhanced MLLM-based VQA framework.}
	\label{fig1}
\end{figure}

Although existing methods~\cite{reflectiva,kurag,mmkbrag,vlmprf} have achieved strong performance in open-domain VQA by integrating external knowledge through multimodal Retrieval-Augmented Generation (RAG) or reorganizing knowledge in a fine-grained manner, they often overlook two critical factors inherent in question-image (QI) pairs, as illustrated in~\cref{fig1}(a): (1) the progressive logical structure implied by the question, and (2) the controllability of the retrieval process. Meanwhile, prior studies~\cite{idealgpt,socratic,chatterbox,pertoken} have shown that Visual Question Decomposition (VQD) and Chain-of-Thought (CoT) reasoning can substantially enhance comprehension and inference in large models. These observations motivate the development of a framework that supports hierarchical question decomposition and logic-guided retrieval for MLLM-based VQA.

\begin{figure}[!t]
	\centering
	\includegraphics[width=0.75\linewidth]{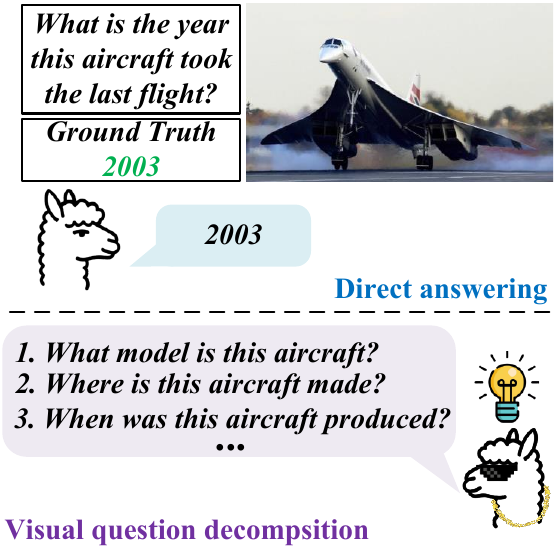}
	\caption{\textbf{The illustration of the VQD.} The generation of Chain-of-Question via VQD on the input question can benefit the MLLM's exploration of the knowledge behind the given image.}
	\label{fig2}
\end{figure}

In this work, we propose a \underline{\textbf{C}}oVQD-\underline{\textbf{g}}uided \underline{\textbf{RAG}} (CgRAG) framework to boost the performance of MLLM-based KBVQA. As illustrated in~\cref{fig1}(b), this framework integrates CoT and VQD into a multi-grained retrieval strategy to achieve the aforementioned performance improvement. Specifically, CgRAG introduces Chain-of-VQD (CoVQD) to extract fine-grained multimodal information from input QI pairs, which is then leveraged to guide a structured RAG process. To effectively incorporate the retrieved knowledge, we design a flexible prompting scheme for MLLM inference.
Moreover, as illustrated in~\cref{fig2}, compelling the MLLM to perform VQD on the input question and output a chain of questions could enhance understanding of knowledge relevant to the given image.
Therefore, to further enhance the analytical capability of MLLMs on QI pairs, we propose a new fine-tuning strategy termed logical implication Direct Preference Optimization (liDPO).

The proposed CgRAG framework consists of three stages: 
1) Dissecting Chain Generation (DCG), which constructs CoT enriched with fine-grained sub-questions for deeper analysis; 
2) Elaborate Knowledge Retrieval (EKR), which leverages DCG outputs to obtain relevant external knowledge; and 
3) Comprehensive Prompt Construction (CPC), which fuses implicit and explicit knowledge into a unified prompt to guide progressive reasoning. 
These stages are sequentially connected, enabling MLLMs to produce more accurate answers accompanied by coherent explanations. Notably, the proposed framework is model-agnostic and can be readily integrated with different MLLMs, thereby unleashing their potential for explanatory VQA.

Our key contributions are summarized in three folds:
\begin{itemize}
 \item We fuse CoT and VQD to construct a logical multi-question chain for knowledge retrieval, and propose a new framework, CoVQD-guided RAG (CgRAG), which enables more abundant and accurate knowledge acquisition to build robust prompts for open-domain VQA with MLLMs.
	
 \item We design a novel fine-tuning strategy for MLLMs, termed liDPO, which further enhances their VQD capability by explicitly encouraging correct logical relations among decomposed sub-questions.
	
 \item Extensive experiments on \textsc{E-VQA}, \textsc{InfoSeek}, and \textsc{OKVQA} demonstrate the effectiveness of the proposed method, achieving competitive performance compared with existing approaches.
\end{itemize}

\section{Related Work}
\label{sec:relate}

\subsection{Knowledge-Based Visual Question Answering}

Knowledge-based Visual Question Answering (KBVQA) requires models to jointly reason about visual content, textual questions, and external knowledge sources to generate accurate answers. According to the modality used during knowledge retrieval, existing retrieval-based KBVQA methods can be broadly categorized into three types: textual-only methods (\emph{e.g.}, Wiki-LLaVA~\cite{wikillava}), visual-only methods (\emph{e.g.}, RORA-VLM~\cite{rora-vlm}, EchoSight~\cite{echosight}, ReflectiVA~\cite{reflectiva}, MMKB-RAG~\cite{mmkbrag}), and methods that combine visual and textual modalities (\emph{e.g.}, DPR\_{V+T}, KU-RAG~\cite{kurag}, VLM-PRF~\cite{vlmprf}). These approaches demonstrate that incorporating external knowledge through retrieval can substantially improve open-domain VQA performance. In contrast to prior work that relies on a single or loosely coupled modality, our work aims to retrieve and exploit fused multimodal information to better support MLLM reasoning. By leveraging hybrid, multi-level knowledge that spans fine-grained visual details, contextual linguistic cues, and external factual resources, the proposed approach facilitates more coherent step-by-step reasoning.

\subsection{Question Decomposition}

Question decomposition has proven effective in enhancing the reasoning capability of LLMs by breaking complex queries into simpler, more manageable sub-questions. Recently, this paradigm has been extended to multimodal scenarios. For instance, You \emph{et al.}~\cite{idealgpt} propose an iterative decomposition framework that enables models to tackle complex vision–language reasoning tasks through progressive steps, while Qi \emph{et al.}~\cite{socratic} demonstrate that Socratic-style recursive questioning can guide LLMs toward deeper and more interpretable reasoning. Inspired by these advances, our work introduces Visual Question Decomposition (VQD) into MLLM-based VQA, enabling complex visual–textual queries to be decomposed into structured sub-questions and to form a chain of questions. This design improves reasoning transparency, enhances retrieval precision, and ultimately leads to more accurate and interpretable answers.

\subsection{Reinforcement Learning for Large Models}

Reinforcement learning (RL)~\cite{rl} is a machine learning paradigm that optimizes its policy to maximize long-term cumulative reward through interaction with the environment, receiving corresponding reward or penalty feedback for its actions. Recently, RL has been widely applied to optimize the behavior of large models and improve their reasoning ability~\cite{deepseekr1,r1onevision}. 
Moreover, with the emergence of Direct Preference Optimization (DPO)~\cite{dpo}, a mainstream RL from Preference Feedback (RLPF)-based preference alignment method, the optimization paradigm can not only preserve the model's inherent generative and inferential capacity but also aligns its output with human-like reasoning norms, factual consistency, and step-by-step logical rigor. 
Building on this line of research, our work develops a DPO strategy that explicitly promotes MLLM's understanding of QI pairs and supports logical question decomposition.

\section{Proposed Method}
\label{sec:metho}

In this section, we elaborate on the proposed CgRAG system, including its overall architecture and key components. An overview of the framework is presented in~\cref{sec:3.1}, highlighting the core design principles. The CgRAG framework is organized into three main components: 
1) the generation of a detailed reasoning chain based on Visual Question Decomposition (VQD), described in \cref{sec:3.2}; 
2) refined Retrieval-Augmented Generation (RAG), detailed in \cref{sec:3.3}; and 
3) comprehensive prompt construction that integrates multi-granular knowledge, outlined in \cref{sec:3.4}.
Together, these components form a cohesive pipeline that effectively enhances open-domain explanatory VQA performance.

\subsection{Overview}
\label{sec:3.1}

Advancing the reasoning capability of LLMs has long been a central goal in artificial intelligence. Recent years have witnessed substantial progress in both unimodal and multimodal LLMs. Although prior studies have improved explanatory VQA through causal reasoning~\cite{xue2023}, contrastive learning~\cite{lai2024}, or the use of frozen LLMs~\cite{xue2024}, the reasoning potential of MLLMs in open-domain explanatory VQA remains insufficiently explored. Existing MLLM-based approaches primarily emphasize enriching fine-grained knowledge through retrieval, while the role of structured question decomposition in guiding retrieval has received less attention. We posit that retrieval explicitly guided by detailed VQD can further improve the effectiveness of reasoning.

\begin{figure*}[!t]
	\centering
	\includegraphics[width=0.9\linewidth]{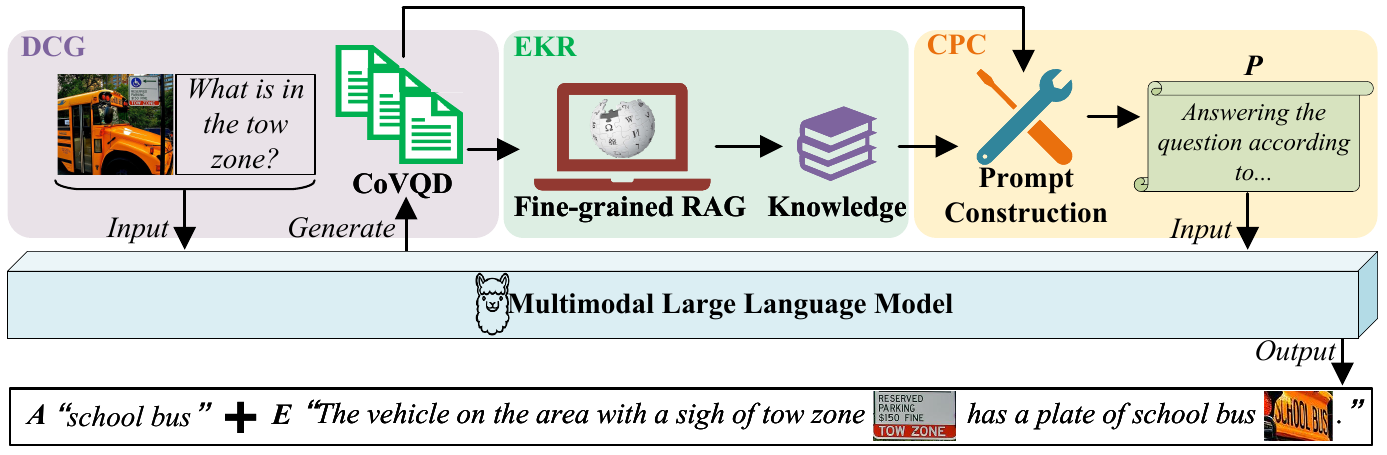}
	\caption{\textbf{Overall architecture of the proposed CgRAG framework.} The pipeline consists of three components: Dissecting Chain Generation (DCG), which constructs CoT guided by VQD from the input image and question; Elaborate Knowledge Retrieval (EKR), which retrieves external knowledge under the guidance of CoVQD; and Comprehensive Prompt Construction (CPC), which aggregates implicit and explicit knowledge for inference.}
	\label{fig3}
\end{figure*}

The proposed CgRAG framework integrates the strengths of existing augmentation strategies by introducing a multi-granular, step-by-step reasoning process based on CoVQD, together with a fine-grained and accurate RAG mechanism. Specifically, to transform the input question into a chain of questions, we fuse CoT reasoning with VQD to form CoVQD, which extracts detailed multimodal information from the input image and question and serves as structured guidance for retrieval. On this basis, we develop an MLLM-based pipeline that can be seamlessly integrated with different backbone MLLMs. A flexible prompt construction strategy is further adopted to effectively organize retrieved knowledge and align it with MLLM inference.

As shown in \cref{fig3}, the proposed framework comprises three stages: Dissecting Chain Generation (DCG), Elaborate Knowledge Retrieval (EKR), and Comprehensive Prompt Construction (CPC). DCG constructs CoT enriched with fine-grained sub-questions derived from VQD, enabling detailed analysis of both visual content and question semantics. EKR subsequently retrieves external knowledge guided by the DCG outputs, providing relevant factual support for reasoning. CPC then integrates the outputs of DCG and EKR into a unified prompt, guiding the MLLM toward progressive and coherent reasoning. Each stage builds upon the previous one, enabling the generation of more accurate answers accompanied by detailed explanations.

\subsection{Dissecting Chain Generation}
\label{sec:3.2}

The DCG module constitutes the first stage of the CgRAG pipeline and integrates CoT reasoning with VQD to produce CoVQD, which plays a central role in subsequent retrieval and reasoning. The predictive mechanism of LLMs in VQA~\cite{pica} and VQD~\cite{khan2023} is rooted in in-context learning~\cite{gpt3}, which enables frozen models to perform zero-shot inference. Given a constructed prompt $p$, a frozen model generates an answer sequence $y=(y_1,\ldots,y_z)$ via autoregressive decoding:
\begin{align}
	\hat{y}^t = \arg \max_{y^t} p_{\theta} \left(y^t \mid p, \hat{y}^{<t} \right),
\end{align}
where $p_{\theta}(\cdot)$ denotes the conditional token distribution parameterized by model weights $\theta$, $\hat{y}^{<t}=\{\hat{y}^1,\ldots,\hat{y}^{t-1}\}$ represents previously generated tokens, and $t$ indexes the decoding step. This objective corresponds to the Next Token Prediction (NTP) loss.

Although existing MLLMs (\emph{e.g.}, LLaVA-1.5~\cite{llava1.5} and Qwen-VL~\cite{qwenvl}) achieve strong performance in VQA, their VQD capability is often limited~\cite{zhang2024}. To address this issue, we first fine-tune the MLLM using a dedicated dataset to improve its VQD ability. Specifically, we adopt the SelectiveVQD loss~\cite{zhang2024}, which combines NTP loss with Binary Cross-Entropy (BCE) loss to supervise question-decomposition decisions. The BCE loss is defined as
\begin{align}
	\mathcal{L}_{\mathrm{BCE}} = -\left[y \log \left(\hat{y} \right) + \left(1-y \right) \log \left(1-\hat{y} \right) \right],
\end{align}
where $y\in\{0,1\}$ is the ground-truth label indicating whether decomposition is required, and $\hat{y}\in(0,1)$ is the predicted probability. The overall SelectiveVQD loss is given by:
\begin{align}
	\mathcal{L}_{\mathrm{SelectiveVQD}} = \sum_{i=1}^{N} \left(\lambda \mathcal{L}_{\mathrm{NTP},i} + \beta \mathcal{L}_{\mathrm{BCE},i} \right),
\end{align}
where $i$ indexes training samples, $N$ is the total number of samples, and $\lambda,\beta$ are weighting hyperparameters balancing the generative and decomposition objectives.

\begin{figure}[!t]
	\centering
	\includegraphics[width = \linewidth]{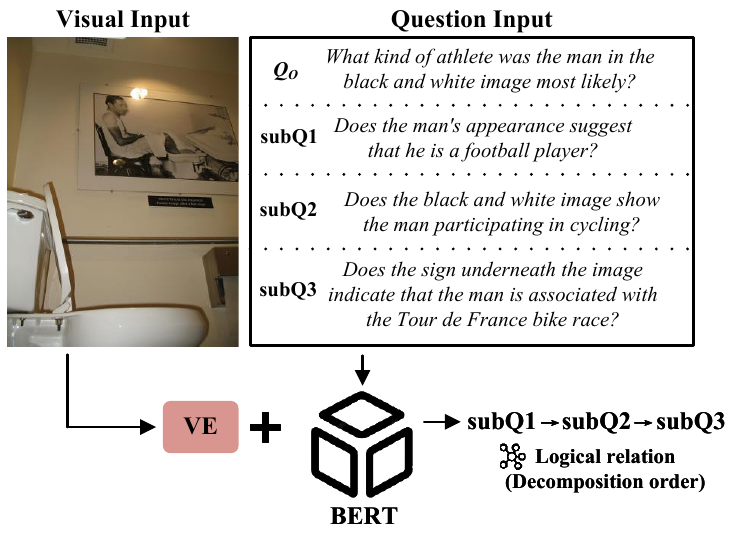}
	\caption{\textbf{Prediction of logical relations among sub-questions.} A pre-trained BERT is employed to infer logical implications between sub-questions for liDPO.}
	\label{fig4}
\end{figure}

\begin{figure*}[!t]
	\centering
	\includegraphics[width=1.0\linewidth]{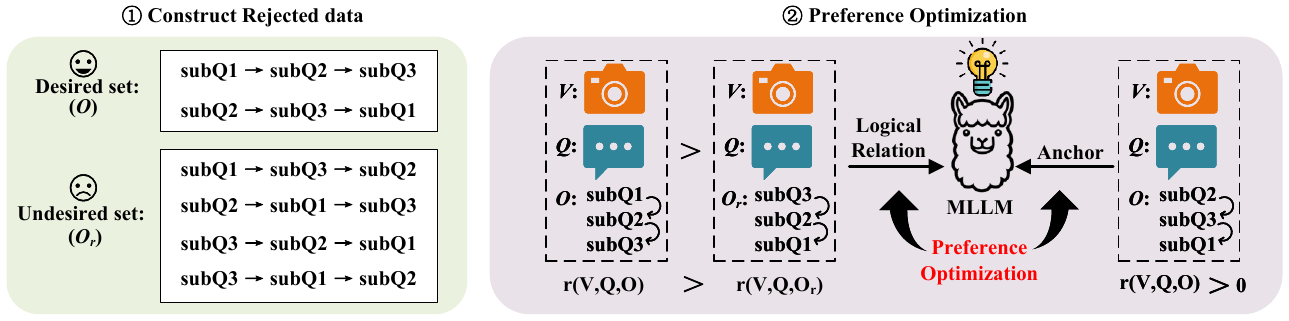}
	\caption{\textbf{Overview of logical implication Direct Preference Optimization (liDPO).} The two-stage procedure includes rejected data construction and preference optimization, where the desired set ($O$) contains logically correct sub-question sequences and the undesired set ($O_r$) contains incorrect ones.}
	\label{fig5}
\end{figure*}

To further enhance the model’s ability to capture logical relations among sub-questions, we introduce logical implication Direct Preference Optimization (liDPO) during fine-tuning. As illustrated in \cref{fig4}, following~\cite{logicalvqa}, we employ a pre-trained BERT~\cite{bert} with a visual encoder to predict logical relations among sub-questions conditioned on both visual and textual inputs. These relations are then used to construct preference pairs for liDPO, as shown in \cref{fig5}. 

In the VQD setting, preference optimization aims to align the model with desired outputs under a reward function while constraining deviation from a reference policy. Given the input question $Q$, image $V$, and an output sub-question sequence $O$, the policy $\pi_{\theta}$ induces a conditional distribution $\pi_{\theta}(O\mid Q,V)$. To avoid over-optimization~\cite{gao2023}, the optimization objective regularizes the divergence between $\pi_{\theta}$ and a reference policy $\pi_{\text{ref}}$ (initialized from the same checkpoint). The corresponding preference-optimization loss is written as:
\begin{align}
	\mathcal{L}_{\mathrm{PO}} = -\log \sigma \left(r \left(Q,V,O \right) - \beta \log \frac{\pi_{\theta} \left(O \mid Q,V \right)}{\pi_{\mathrm{ref}} \left(O \mid Q,V \right)} \right),
\end{align}
where $r(Q,V,O)$ denotes the reward associated with $O$, $\sigma(\cdot)$ is the sigmoid function, and $\beta$ controls the regularization strength.

Direct Preference Optimization (DPO)~\cite{dpo} further simplifies this alignment by directly maximizing the reward gap between a preferred output $O_w$ and a rejected output $O_l$ under the Bradley-Terry model~\cite{bradley1952}. Its objective is the following.
\begin{align}
	\mathcal{L}_{\mathrm{DPO}} = -\log \sigma \bigg(
	& \beta \log \frac{\pi_{\theta} \left(O_w \mid Q,V \right)}{\pi_{\mathrm{ref}} \left(O_w \mid Q,V \right)} \nonumber \\
	- & \beta \log \frac{\pi_{\theta} \left(O_l \mid Q,V \right)}{\pi_{\mathrm{ref}} \left(O_l \mid Q,V \right)}
	\bigg),
\end{align}
where $O_w$ and $O_l$ are the preferred and rejected sub-question sequences, respectively. 

As a multimodal optimization objective, the proposed liDPO is implemented as a two-stage procedure. In the first stage, we construct preference pairs using the logical relations predicted by BERT: the sub-question order consistent with the desired logical relation serves as $O_w$, while inconsistent orders are treated as $O_l$. In the second stage, we optimize the MLLM with the DPO objective. Moreover, inspired by~\cite{mdpo,odpo}, we incorporate an anchor-based objective to consistently reinforce high-quality outputs:
\begin{align}
	\mathcal{L}_{\mathrm{AncPO}} = -\log \sigma \left(
	\beta \log \frac{\pi_{\theta} \left(O \mid Q,V \right)}{\pi_{\mathrm{ref}} \left(O \mid Q,V \right)}
	\right),
\end{align}
where $O$ denotes an anchor output with high quality. The final liDPO loss is:
\begin{align}
	\mathcal{L}_{\mathrm{liDPO}} = \mathcal{L}_{\mathrm{DPO}} + \gamma \mathcal{L}_{\mathrm{AncPO}},
\end{align}
where $\gamma$ weights the anchored objective.

During inference, given an image-question pair $(I_O,Q_O)$, the fine-tuned MLLM decomposes $Q_O$ into an ordered set of sub-question-answer pairs $\{qa_1,\ldots,qa_n\}$, forming a structured CoVQD. Here, $n$ denotes the number of decomposed sub-questions. This structure provides interpretable intermediate reasoning steps and serves as explicit guidance for downstream retrieval.

Formally, the generated CoVQD is organized as an ordered chain that preserves both the visual context and the sequential dependency among sub-question-answer pairs:

\textbf{Head:} Please decompose the given question into sub-questions for easier answering according to the given image.

\textbf{Context: $I_O$}

\textbf{Question: $Q_O$}

\begin{align}
	\begin{array}{c}
	\boxed{I_O \quad \setminus n \quad qa_1 \quad \setminus n \quad \ldots \quad \setminus n \quad qa_n}
	\end{array}
\end{align}
where $\setminus n$ denotes a delimiter separating consecutive elements in the chain. This explicit structure serves as a unified intermediate representation that is subsequently used to guide fine-grained knowledge retrieval during the EKR stage.

\begin{figure*}[!t]
	\centering
	\includegraphics[width=1.0\linewidth]{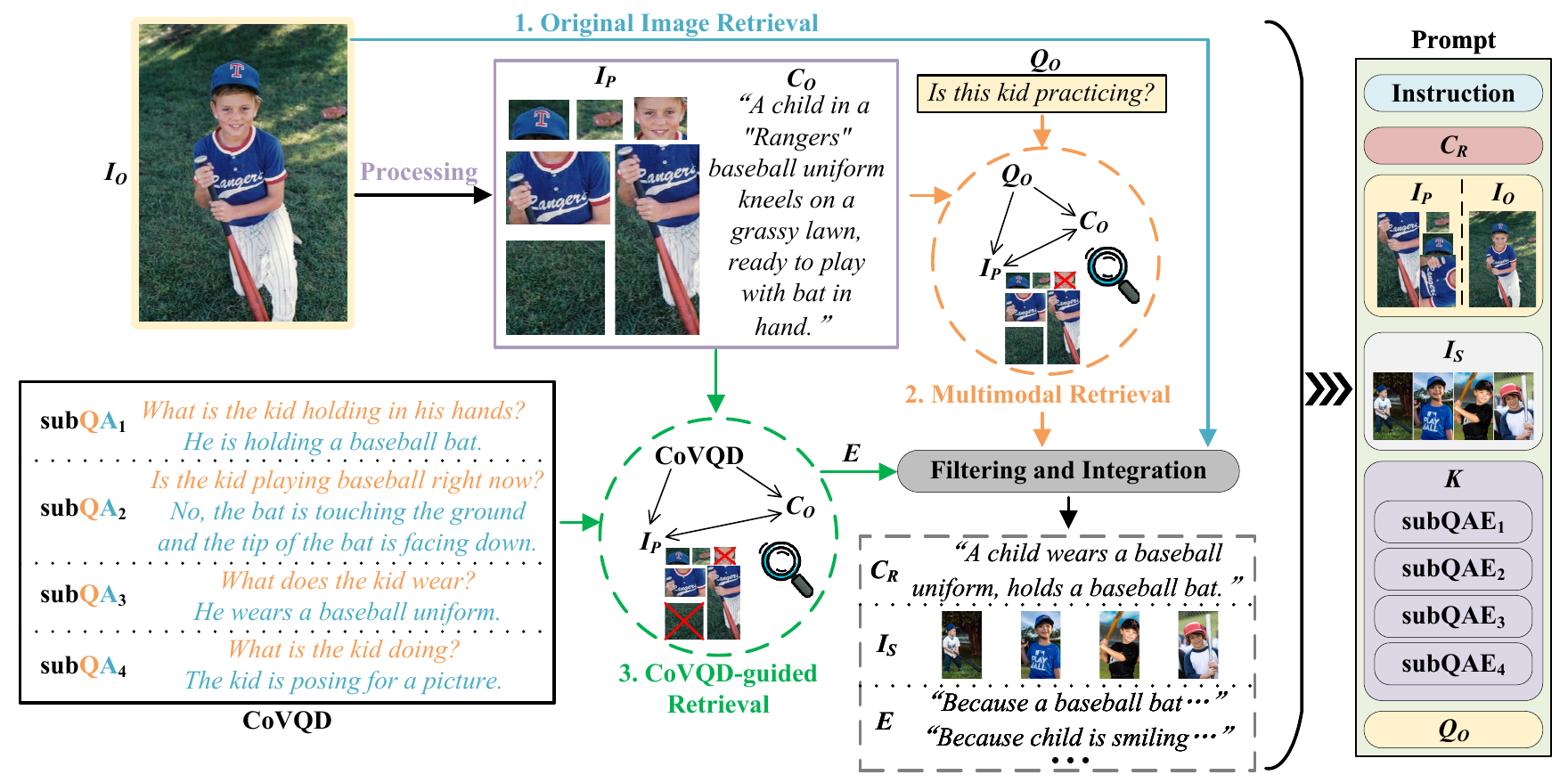}
	\caption{\textbf{Illustration of the Elaborate Knowledge Retrieval (EKR) module.} Three retrieval processes are involved: original image retrieval, multimodal retrieval, and CoVQD-guided retrieval. $I$, $C$, $E$, and $K$ denote Image, Caption, Explanation, and Knowledge, respectively. And $Q_O$ and CoVQD serve as a supervisor for filtering out visual information irrelevant to reasoning.}
	\label{fig6}
\end{figure*}

\subsection{Elaborate Knowledge Retrieval}
\label{sec:3.3}
 
Following DCG, the EKR module performs multi-level external knowledge retrieval. As illustrated in \cref{fig6}, it consists of three retrieval processes: original image retrieval, multimodal retrieval, and CoVQD-guided retrieval, corresponding to increasing levels of granularity. The original image $I_O$, the question $Q_O$, and the generated CoVQD jointly serve as retrieval inputs. It is worth noting that in multimodal retrieval and CoVQD-guided retrieval, $Q_O$ and CoVQD act as a supervisor for filtering out visual information irrelevant to reasoning, respectively.
After the three retrieval and the filtering method following~\cite{vlmprf}, the more comprehensive and fine-grained multimodal information is obtained, which contains the Refined Caption $C_R$, Searched Image $I_S$, and Explanation $E$. And this information would be integrated into the final prompt for the MLLM's prediction. 

Concretely, original image retrieval provides coarse visual context. Multimodal retrieval leverages VinVL~\cite{vinvl} and BLIP-2~\cite{blip2} to generate refined image patches $I_P$ and a global caption $C_O$, which are retrieved under the supervision of $Q_O$. CoVQD-guided retrieval further refines this process by sequentially conditioning retrieval on each sub-question-answer pair, ensuring logical consistency across $n$ retrieval steps. Retrieved visual and textual candidates are projected into a shared embedding space using vision-language encoders such as CLIP~\cite{clip} or BLIP~\cite{blip}. For each sub-question $(q_i,a_i)$, cosine similarity is computed to select the most aligned explanation $e_i$, yielding a set of Question-Answer-Explanation (QAE) triples $\{(q_i,a_i,e_i)\}_{i=1}^{n}$.

\subsection{Comprehensive Prompt Construction}
\label{sec:3.4}

The CPC stage integrates the outputs of DCG and EKR to guide MLLM inference. Unlike existing approaches, our prompt jointly incorporates refined visual content and structured logical knowledge, reducing irrelevant information and mitigating hallucinations. The constructed prompt consists of five components: instruction head $H$, refined caption $C_R$, image patches $I_P$, logical knowledge $K=\{(q_i,a_i,e_i)\}_{i=1}^{n}$, and the original question $Q_O$:
\begin{align}
	\begin{array}{c}
	\boxed{H \setminus n C_R \setminus n I_P \setminus n K \setminus n Q_O}.
	\end{array}
\end{align}
By integrating these components into a unified prompt, the proposed CgRAG framework provides a comprehensive, multi-granular context for inference, improving answer accuracy and explanation reliability in open-domain explanatory VQA.

\section{Experimental Results}
\label{sec:experi}

In this section, we design explicit experiments to validate the effectiveness of the proposed CgRAG. Specifically, the experimental settings and research questions are presented in \cref{sec:4.1}; the main quantitative comparisons against existing methods and evaluations with different backbone models are reported in \cref{sec:4.2}; exploratory studies that analyze key components and other influential factors are provided in \cref{sec:4.3}; and a qualitative analysis of CgRAG’s out-of-domain (OOD) behavior is presented in \cref{sec:4.4}.

\subsection{Experimental Settings}
\label{sec:4.1}

\subsubsection{Datasets and Metrics} 

To evaluate the reasoning capacity of CgRAG in zero-shot open-ended KBVQA, we adopt \textsc{OK-VQA}~\cite{ok-vqa}, \textsc{Encyclopedic-VQA (E-VQA)}~\cite{encyvqa}, and \textsc{InfoSeek}~\cite{infoseek} in the main experiments. Specifically, \textsc{OK-VQA} is a widely used KBVQA benchmark containing 9K and 5K image-question pairs for training and testing, respectively. \textsc{E-VQA} is a large-scale dataset that focuses on fine-grained categories and instance-level properties. It contains approximately 221K unique question-answer pairs, each associated with up to five images, resulting in nearly 1M VQA samples. Moreover, \textsc{E-VQA} provides a controlled knowledge base derived from Wikipedia, which supplies explicit evidence for each annotated answer and enables knowledge-grounded evaluation. \textsc{InfoSeek} is a large-scale benchmark for information-seeking VQA, in which answering requires external knowledge beyond common sense. It comprises approximately 1.3M VQA pairs aligned with 11K images sourced from OVEN~\cite{oven}. The data are split into a training set (934K pairs) and a validation set (73K pairs), with strict separation by both entities and questions. The validation set is further divided into Unseen Entity and Unseen Question subsets, facilitating a fine-grained evaluation of generalization to novel concepts and query forms. In addition, we employ GQA-REX~\cite{rex}, which extends the GQA benchmark~\cite{gqa} with multimodal annotations that capture the visual reasoning process, and we evaluate on GQA-OOD~\cite{gqa-ood}, a benchmark specifically designed for assessing OOD robustness.


For VQA, we use VQA accuracy~\cite{goyal2017} to evaluate answer correctness. For explanation generation, we adopt standard automatic metrics, including BLEU-4, METEOR, ROUGE-L, CIDEr, and SPICE, to assess explanation quality, and we use the Grounding metric~\cite{rex} to evaluate whether the generated explanation correctly localizes relevant visual regions.

\subsubsection{Methods and MLLMs} 

To assess the proposed method objectively, we compare two categories of MLLM-based VQA approaches: zero-shot MLLMs and retrieval-augmented models. For zero-shot MLLMs, we evaluate BLIP-2~\cite{blip2}, InstructBLIP~\cite{instructblip}, LLaVA-1.5~\cite{llava1.5}, GPT-4V~\cite{gpt4v}, Qwen2-VL~\cite{qwen2vl}, Qwen2.5-VL~\cite{qwen2.5vl}, and LLaVA-NeXT~\cite{llavanext}. For retrieval-augmented models, we compare against DPR\_{V+T}~\cite{dprv+t}, RORA-VLM~\cite{rora-vlm}, EchoSight~\cite{echosight}, Wiki-LLaVA~\cite{wikillava}, ReflectiVA~\cite{reflectiva}, KU-RAG~\cite{kurag}, MMKB-RAG~\cite{mmkbrag}, and VLM-PRF~\cite{vlmprf}. 
To explore generalization within the same pipeline, we further instantiate CgRAG with different backbone MLLMs (Qwen2-VL, Qwen2.5-VL, LLaVA-NeXT, and InternVL3~\cite{internvl3}).

\subsubsection{Implementation} 

For MLLM fine-tuning, we follow the training strategy of~\cite{zhang2024}: all MLLMs are fine-tuned on DecoVQD+ using the SelectiveVQD loss. For BERT pretraining, we follow~\cite{logicalvqa}: BERT is pretrained on \textsc{SNLI}~\cite{snli} for five epochs, initialized with \emph{bert-base-uncased}, using a batch size of 16, weight decay of 0.01, and AdamW with a learning rate of $2\times10^{-5}$. The same configuration is then used to fine-tune the model on a subset of 2,000 manually annotated proposition pairs from \textsc{Introspect}~\cite{introspect}. For liDPO fine-tuning, we set $\beta$ as 0.5, and the learning rate is set up to $1\times10^{-7}$, employ a cosine scheduler with a warmup ratio of 0.03, and set $\gamma=1$ by default. All MLLMs are trained for one epoch.

\subsubsection{Research Question}

To evaluate CgRAG comprehensively, we design experiments around four research questions (RQs): 
1) assessing the effectiveness of CgRAG relative to MLLM-based methods in zero- and few-shot settings; 
2) testing the generalization of CgRAG across different backbone MLLMs; 
3) quantifying the contribution of each component and analyzing the impact of key factors; and 
4) qualitatively assessing robustness across diverse domains. The RQs are summarized as follows:
\begin{itemize}
	\item \emph{RQ\_1}: How does CgRAG compare fairly with zero-shot and few-shot methods?
	\item \emph{RQ\_2}: How does CgRAG’s performance vary across different backbone MLLMs within the same pipeline?
	\item \emph{RQ\_3}: What are the contributions of each component, and how do other factors influence performance?
	\item \emph{RQ\_4}: How does CgRAG perform across evaluation cases from different domains?
\end{itemize}
The corresponding experiments are presented in the following sections.

\begin{table*}[!t]
	\centering
	\caption{\textbf{Overall performance comparison on \textsc{E-VQA} and \textsc{InfoSeek} datasets.} Bold and underlined values indicate the best and second-best results, respectively. ``V'' and ``T'' denote visual and textual features.}
	\setlength\tabcolsep{4pt}
	\begin{tabular}{lllc|cc|ccc}
	\toprule[1.1pt]
	\multirow{2}[2]{*}{{Method}} & \multirow{2}[2]{*}{{LLM/MLLM}} & \multirow{2}[2]{*}{{Retriever}} & \multirow{2}[2]{*}{{Feature}} & \multicolumn{2}{c|}{\textsc{{E-VQA}}} & \multicolumn{3}{c}{\textsc{{InfoSeek}}} \\
	\cmidrule(lr){5-6} \cmidrule(lr){7-9}
	& & & & {Single-Hop} & {All} & {Unseen-Q} & {Unseen-E} & {All} \\
	\midrule
	\multicolumn{9}{c}{\emph{Zero-shot MLLMs}} \\
	\midrule
	InstructBLIP~\cite{instructblip} {\smaller (NeurIPS'23)} & Flan-$T5_\mathrm{XL}$ & - & - & 11.9 & 12.0 & 8.9 & 7.4 & 8.1 \\
	BLIP-2~\cite{blip2} {\smaller (ICML'23)} & Flan-$T5_\mathrm{XL}$ & - & - & 12.6 & 12.4 & 12.7 & 12.3 & 12.5 \\
	GPT-4V~\cite{gpt4v} {\smaller (arxiv'24)} & - & - & - & 26.8 & 28.0 & 15.0 & 14.3 & 14.6 \\
	LLaVA-1.5~\cite{llava1.5} {\smaller (CVPR'24)} & LLaMA-3.1-8B & - & - & 16.0 & 16.9 & 8.3 & 8.9 & 7.8 \\
	LLaVA-1.5~\cite{llava1.5} {\smaller (CVPR'24)} & Vicuna-7B & - & - & 16.3 & 16.9 & 9.6 & 9.4 & 9.5 \\
	LLaVA-NeXT-7B~\cite{llavanext} {\smaller (arxiv'24)} & - & - & - & 22.1 & 20.8 & 23.2 & 23.4 & 23.1 \\
	LLaVA-NeXT-8B~\cite{llavanext} {\smaller (arxiv'24)} & - & - & - & 22.5 & 21.4 & 23.7 & 24.0 & 23.6 \\ 
	Qwen2-VL-7B~\cite{qwen2vl} {\smaller (arxiv'24)} & - & - & - & 16.2 & 17.0 & 15.4 & 16.8 & 16.1 \\
	Qwen2.5-VL-3B~\cite{qwen2.5vl} {\smaller (arxiv'25)} & - & - & - & 17.9 & 19.6 & 20.2 & 21.7 & 20.7 \\
	Qwen2.5-VL-7B~\cite{qwen2.5vl} {\smaller (arxiv'25)} & - & - & - & 21.7 & 20.3 & 22.8 & 24.0 & 23.2 \\
	\midrule
	\multicolumn{9}{c}{\emph{Retrieval-Augmented Models}} \\
	\midrule
	Wiki-LLaVA~\cite{wikillava} {\smaller (CVPRW'24)} & LLaMA-3.1-8B & CLIP ViT-L/14 + Contriever & T & 18.3 & 19.6 & 28.6 & 25.7 & 27.1 \\
	Wiki-LLaVA~\cite{wikillava} {\smaller (CVPRW'24)} & Vicuna-7B & CLIP ViT-L/14 + Contriever & T & 17.7 & 20.3 & 30.1 & 27.8 & 28.9 \\
	RORA-VLM~\cite{rora-vlm} {\smaller (arxiv'24)} & Vicuna-7B & CLIP + Google Search & V & - & 20.3 & 25.1 & 27.3 & 26.0 \\
	EchoSight~\cite{echosight} {\smaller (EMNLP'24)} & Mistral-7B/LLaMA-3-8B & EVA-CLIP-8B & V & 19.4 & - & - & - & 27.5 \\
	EchoSight~\cite{echosight} {\smaller (EMNLP'24)} & LLaMA-3.1-8B & EVA-CLIP-8B & V & 26.4 & 24.9 & 18.0 & 19.8 & 18.8 \\
	ReflectiVA~\cite{reflectiva} {\smaller (CVPR'25)} & LLaMA-3.1-8B & EVA-CLIP-8B & V & 35.5 & 35.5 & 28.6 & 28.1 & 28.3 \\
	MMKB-RAG~\cite{mmkbrag} {\smaller (arxiv'25)} & Qwen2-VL-7B & EVA-CLIP-8B & V & 39.7 & 35.9 & 36.4 & 36.3 & 36.4 \\
	\midrule
	DPR\_{V+T}~\cite{dprv+t} {\smaller (ECIR'24)} & Multi-passage BERT & CLIP ViT-B/32 & V+T & 29.1 & - & - & - & 12.4 \\
	KU-RAG~\cite{kurag} {\smaller (arxiv'25)} & GPT-4o & EVA-CLIP-8B & V+T & 38.3 & - & - & - & 26.1 \\
	VLM-PRF~\cite{vlmprf} {\smaller (NeurIPS'25)} & Qwen2.5-VL-7B & EVA-CLIP-8B & V+T & 28.9 & 28.6 & 40.0 & 39.4 & 39.5 \\
	VLM-PRF \emph{w/} RL~\cite{vlmprf} {\smaller (NeurIPS'25)} & InternVL3-8B & EVA-CLIP-8B & V+T & \underline{40.1} & \underline{39.2} & \textbf{43.5} & \textbf{42.1} & \underline{42.5} \\
	\rowcolor{gray!20}
	{CgRAG (Ours)} & Qwen2-VL-7B & EVA-CLIP-8B & V+T & 39.6 & 38.3 & 40.8 & 39.2 & 36.3 \\
	\rowcolor{gray!20}
	{CgRAG (Ours)} & Qwen2.5-VL-7B & EVA-CLIP-8B & V+T & 39.8 & 38.6 & 41.3 & 39.5 & 36.6 \\
	\rowcolor{gray!20}
	{CgRAG (Ours)} & LLaVA-NeXT-7B & EVA-CLIP-8B & V+T & 39.9 & 39.0 & 41.7 & 40.5 & 39.5 \\
	\rowcolor{gray!20}
	{CgRAG (Ours)} & LLaVA-NeXT-8B & EVA-CLIP-8B & V+T & 39.9 & \underline{39.2} & 41.9 & 40.9 & 39.7 \\
	\rowcolor{gray!20}
	{CgRAG (Ours)} & InternVL3-8B & EVA-CLIP-8B & V+T & \textbf{40.4} & \textbf{39.5} & \textbf{43.5} & \underline{42.0} & \textbf{43.0} \\
	\bottomrule[1.1pt]
	\end{tabular}
	\label{tab1}%
\end{table*}

\begin{figure}[!t]
	\centering
	\includegraphics[width = 0.9\linewidth]{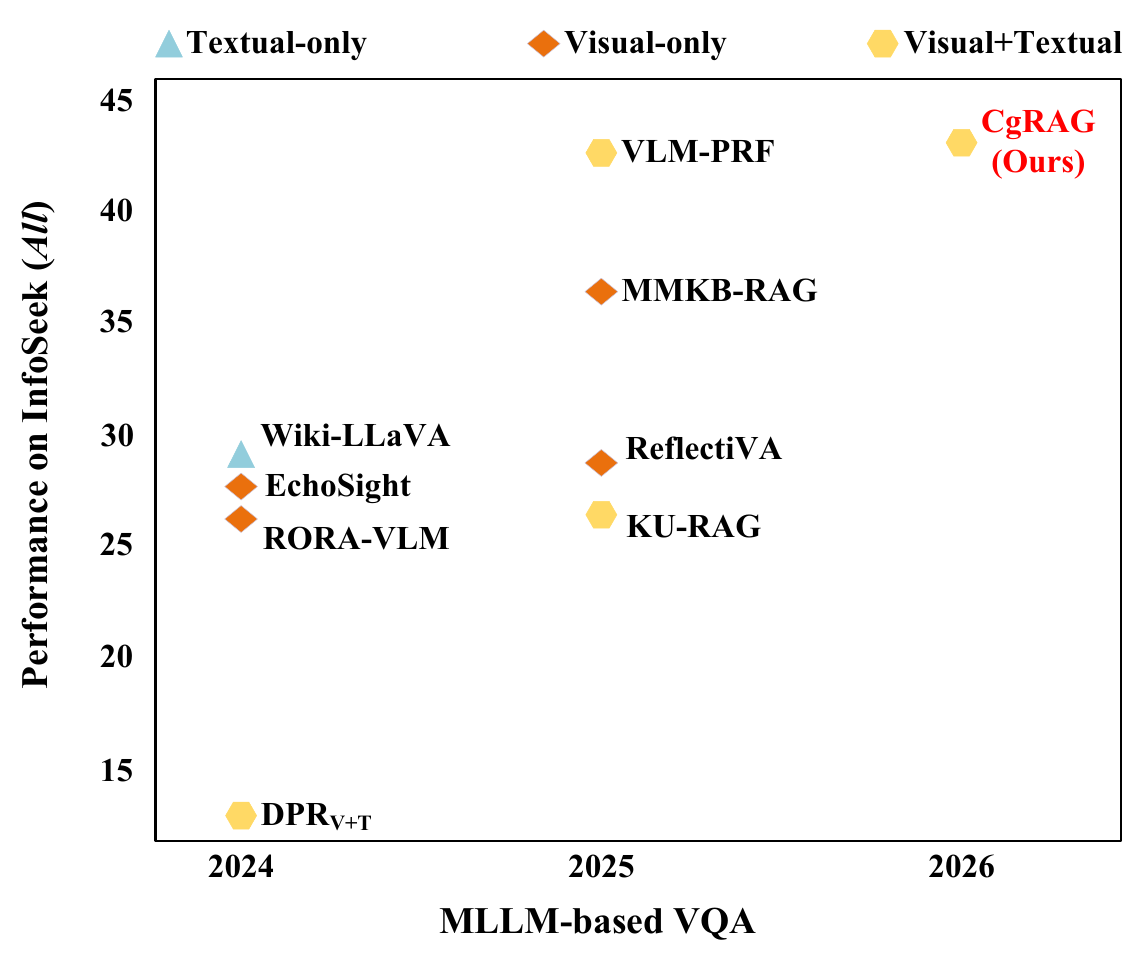}
	\caption{\textbf{Performance comparison of MLLM-based VQA on \textsc{InfoSeek}-\emph{All}.} The best result of each method is reported, and different shapes indicate distinct retrieval features.}
	\label{fig7}
\end{figure}

\begin{table}[!t]
	\centering
	\caption{\textbf{Performance comparison on \textsc{OK-VQA} dataset.} Bold values indicate the best results.}
	\setlength\tabcolsep{12pt}
	\begin{tabular}{l|cc}
	\toprule[1.1pt]
	{Method} & {LLM/MLLM} & {Accuracy} \\
	\midrule
	Qwen2.5-VL-3B~\cite{qwen2.5vl} & - & 62.0 \\
	Qwen2.5-VL-7B~\cite{qwen2.5vl} & - & 72.4 \\
	KU-RAG~\cite{kurag} & GPT-4o & 77.2 \\
	\rowcolor{gray!20}
	{CgRAG (Ours)} & InternVL3-8B & \textbf{77.8} \\
	\bottomrule[1.1pt]
	\end{tabular}
	\label{tab2}
\end{table}

\subsection{Quantitative Results}
\label{sec:4.2}

To address \emph{RQ\_1} and \emph{RQ\_2}, we compare CgRAG with two categories of MLLM-based VQA methods: zero-shot MLLMs and retrieval-augmented models. We instantiate CgRAG with multiple backbone MLLMs to examine its generalization within a unified pipeline. \cref{tab1} reports results on \textsc{E-VQA} and \textsc{InfoSeek}, and \cref{fig7} provides an intuitive comparison on \textsc{InfoSeek}-\emph{All}. Overall, CgRAG with InternVL3~\cite{internvl3} achieves a new state-of-the-art on \textsc{E-VQA} and \textsc{InfoSeek} (\emph{Unseen-Q} and \emph{All}), while CgRAG paired with other MLLMs also yields strong accuracy.

Specifically, CgRAG with InternVL3 attains 40.4\% and 39.5\% on \emph{Single-Hop} and \emph{All} of \textsc{E-VQA}, and 43.5\% and 43.0\% on \emph{Unseen-Q} and \emph{All} of \textsc{InfoSeek}, which are the best results among all compared methods. It achieves 42.0\% on \emph{Unseen-E} of \textsc{InfoSeek}, which is slightly below the best-performing model. Compared with methods that use both visual and textual features, CgRAG improves over DPR\_{V+T} by 11.3\% on \emph{Single-Hop} of \textsc{E-VQA} and by 30.6\% on \textsc{InfoSeek}; surpasses KU-RAG by 2.1\% on \emph{Single-Hop} of \textsc{E-VQA} and by 16.9\% on \textsc{InfoSeek}; and exceeds VLM-PRF by 11.5\% and 10.9\% on \emph{Single-Hop} and \emph{All} of \textsc{E-VQA}, and by 3.5\%, 2.6\%, and 3.5\% on \emph{Unseen-Q}, \emph{Unseen-E}, and \emph{All} of \textsc{InfoSeek}, respectively. Compared with the latest VLM-PRF \emph{w/} RL, CgRAG achieves slightly higher accuracy on \emph{Single-Hop} and \emph{All} of \textsc{E-VQA} (both +0.3\%) and on \emph{All} of \textsc{InfoSeek} (+0.5\%), while being marginally lower on \emph{Unseen-E} of \textsc{InfoSeek} (-0.1\%).

These results indicate that CoVQD-guided refined retrieval effectively strengthens MLLMs for open-domain KBVQA by improving grounding and maintaining semantic coherence across modalities. By using a chain-of-questions as a structured retrieval signal, the model is encouraged to retrieve and integrate evidence aligned with the decomposed reasoning steps, thereby enabling robust cross-modal inference on knowledge-intensive questions.

To further validate CgRAG for open-ended KBVQA, we compare it with existing methods on \textsc{OK-VQA}. The results are summarized in \cref{tab2}. CgRAG consistently outperforms prior retrieval-augmented baselines, and CgRAG with InternVL3 achieves the best performance among the MLLM-based methods compared. This demonstrates that CgRAG effectively combines multimodal reasoning with external knowledge retrieval, supporting accurate answers to knowledge-intensive questions beyond surface-level visual recognition.

\begin{table*}[!t]
	\centering
	\caption{\textbf{Results on explanation generation and question answering for explanatory VQA.} GQA- and OOD- denote answer accuracy on \textsc{GQA-REX} and \textsc{GQA-OOD}, respectively. Bold values indicate the best results.}
	\begin{tabular}{l|cccccc|cccc}
	\toprule[1.1pt]
	{Method} & {BLEU-4} & {METEOR} & {ROUGE-L} & {CIDEr} & {SPICE} & {Grounding} & {GQA-val} & {GQA-test} & {OOD-val} & {OOD-test} \\
	\midrule
	VQA-E~\cite{vqa-e} \smaller{(ECCV'18)} & 42.5 & 34.5 & 73.5 & 358.1 & 40.3 & 31.2 & 65.1 & 57.2 & 49.1 & 46.2 \\
	EXP~\cite{exp} \smaller{(ACLW'19)} & 42.4 & 34.4 & 73.4 & 357.0 & 40.3 & 33.5 & 65.1 & 56.9 & 49.4 & 47.6 \\
	REX~\cite{rex} \smaller{(CVPR'22)} & 54.7 & 39.5 & 79.3 & 465.9 & 49.9 & 70.7 & 78.1 & 58.1 & 71.2 & 52.1 \\
	VCIN~\cite{vcin} \smaller{(ICCV'23)} & 58.6 & 41.5 & 81.4 & 519.2 & 54.6 & 77.3 & 81.7 & 60.6 & 74.7 & 54.2 \\
	MRVQA~\cite{mrvqa} \smaller{(CVPRW'25)} & 16.9 & 23.0 & 44.2 & 107.0 & 23.6 & 18.4 & 38.2 & 32.7 & 28.0 & 26.5 \\
	\rowcolor{gray!20}
	{CgRAG (Ours)} & \textbf{61.4} & \textbf{44.1} & \textbf{83.8} & \textbf{590.2} & \textbf{57.9} & \textbf{82.3} & \textbf{84.3} & \textbf{64.4} & \textbf{78.0} & \textbf{57.5} \\
	\bottomrule[1.1pt]
	\end{tabular}
	\label{tab3}
\end{table*}

\begin{figure}[!t]
	\centering
	\includegraphics[width = 0.9\linewidth]{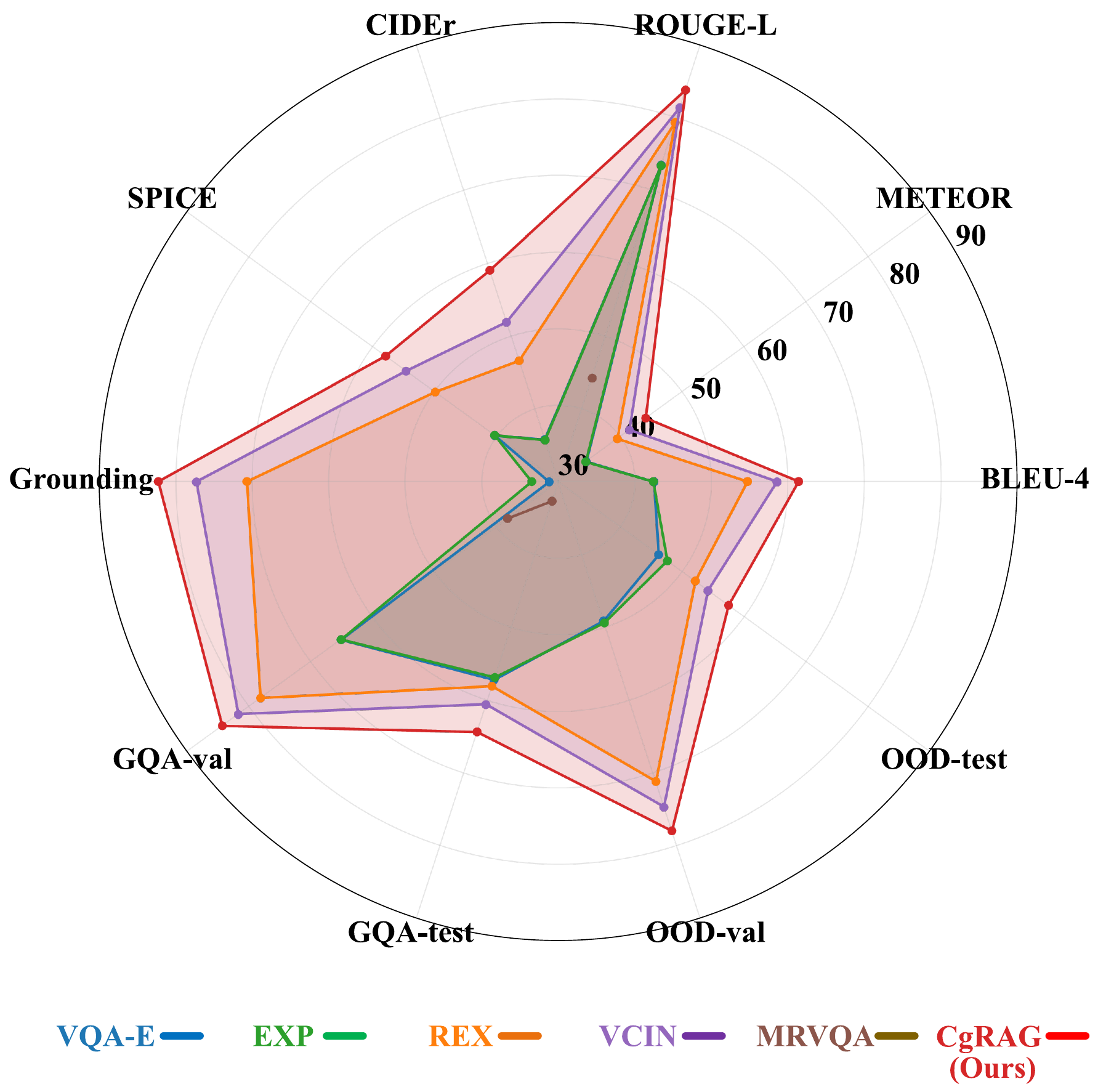}
	\caption{\textbf{Comparison of different methods on Explanatory VQA.} For visualization clarity, CIDEr scores are scaled to one-tenth of their original values.}
	\label{fig8}
\end{figure}

\begin{table}[!t]
	\centering
	\caption{\textbf{Consistency evaluation between predicted answers and explanations on \textsc{GQA-REX}.} ``Con.'', ``Vis.'', and ``Tex.'' denote consistency, visual relevance, and textual relevance, respectively. Bold values indicate the best results.}
	\setlength\tabcolsep{12pt}
	\begin{tabular}{l|cccc}
	\toprule[1.1pt]
	{Method} & {Con.} & {Vis.} & {Tex.} & {Ave.} \\
	\midrule
	REX~\cite{rex} & 84.8 & 3.1 & 4.1 & 3.6 \\
	VCIN~\cite{vcin} & 93.4 & 3.5 & 4.5 & 3.9 \\
	\rowcolor{gray!20}
	{CgRAG (Ours)} & \textbf{97.0} & \textbf{3.9} & \textbf{4.9} & \textbf{4.5} \\
	\bottomrule[1.1pt]
	\end{tabular}
	\label{tab4}
\end{table}

\begin{table}[!t]
	\centering
	\caption{\textbf{Ablation study of DCG, EKR, and CPC components in the CgRAG framework.} Bold values indicate the best results.}
	\setlength\tabcolsep{7pt}
	\begin{tabular}{ccc|ccc}
	\toprule[1.1pt]
	{DCG} & {EKR} & {CPC} & \textsc{{E-VQA}} (Single-Hop) & \textsc{{InfoSeek}} (All) \\
	\midrule
	\Circle & \Circle & \Circle & 23.1 & 24.0 \\
	\CIRCLE & \Circle & \Circle & 29.6 & 30.3 \\
	\Circle & \CIRCLE & \Circle & 37.5 & 35.4 \\
	\CIRCLE & \CIRCLE & \Circle & 39.6 & 36.1 \\
	\Circle & \CIRCLE & \CIRCLE & 38.0 & 35.9 \\
	\rowcolor{gray!20}
	\CIRCLE & \CIRCLE & \CIRCLE & \textbf{40.4} & \textbf{43.0} \\
	\bottomrule[1.1pt]
	\end{tabular}
	\label{tab5}
\end{table}

In addition, to evaluate the quality of explanations produced within CgRAG, we adapt the framework to generate multimodal explanations and assess performance on \textsc{GQA-REX}. Following Chen and Zhao~\cite{rex}, we evaluate both VQA accuracy and explanation quality. 
To evaluate answer-explanation consistency, we additionally report Visual Consistency (Vis.) and Textual Consistency (Tex.)~\cite{vcin}. \cref{tab3,fig8} shows comparisons with VQA-E~\cite{vqa-e}, EXP~\cite{exp}, REX~\cite{rex}, VCIN~\cite{vcin}, and MRVQA~\cite{mrvqa}. Overall, CgRAG achieves the best scores across all explanation metrics, indicating that it produces fluent, informative, and well-grounded explanations. In terms of question answering, CgRAG also achieves the highest accuracy on both in-distribution (\textsc{GQA-REX} validation/test) and out-of-distribution (\textsc{GQA-OOD} validation/test) splits, demonstrating robustness and generalization.

\begin{table*}[!t]
	\centering
	\caption{\textbf{Ablation study of different retrieval modes within the CgRAG framework.} Bold and underlined values indicate the best and second-best results, ``V'' and ``T'' denote visual and textual features, respectively. And their results are highlighted in different colors.}
	\setlength\tabcolsep{12pt}
	\begin{tabular}{l|c|cc|ccc}
	\toprule[1.1pt]
	\multirow{2}[2]{*}{{MLLM}} & \multirow{2}[2]{*}{{Feature}} & \multicolumn{2}{c|}{\textsc{{E-VQA}}} & \multicolumn{3}{c}{\textsc{{InfoSeek}}} \\
	\cmidrule(lr){3-4} \cmidrule(lr){5-7}
	& & {Single-Hop} & {All} & {Unseen-Q} & {Unseen-E} & {All} \\
	\midrule
	\multirow{3}{*}{Qwen2-VL-7B~\cite{qwen2vl} \smaller{(arxiv'24)}} & \lightbluecell{T} & \lightbluecell{22.1} & \lightbluecell{23.4} & \lightbluecell{25.6} & \lightbluecell{25.5} & \lightbluecell{25.5} \\
	& \palepinkcell{V} & \palepinkcell{37.8} & \palepinkcell{36.2} & \palepinkcell{34.9} & \palepinkcell{34.6} & \palepinkcell{34.7} \\
	& \orangecell{V+T} & \orangecell{39.6} & \orangecell{38.3} & \orangecell{40.8} & \orangecell{39.2} & \orangecell{36.3} \\
	\midrule
	\multirow{3}{*}{Qwen2.5-VL-7B~\cite{qwen2.5vl} \smaller{(arxiv'25)}} & \lightbluecell{T} & \lightbluecell{22.6} & \lightbluecell{24.0} & \lightbluecell{27.7} & \lightbluecell{27.0} & \lightbluecell{27.3} \\
	& \palepinkcell{V} & \palepinkcell{38.3} & \palepinkcell{36.9} & \palepinkcell{35.5} & \palepinkcell{35.7} & \palepinkcell{35.5} \\
	& \orangecell{V+T} & \orangecell{39.8} & \orangecell{38.6} & \orangecell{41.3} & \orangecell{39.5} & \orangecell{36.6} \\
	\midrule
	\multirow{3}{*}{LLaVA-NeXT-7B~\cite{llavanext} \smaller{(arxiv'24)}} & \lightbluecell{T} & \lightbluecell{22.3} & \lightbluecell{23.6} & \lightbluecell{27.2} & \lightbluecell{26.3} & \lightbluecell{26.9} \\
	& \palepinkcell{V} & \palepinkcell{37.4} & \palepinkcell{35.6} & \palepinkcell{33.8} & \palepinkcell{34.5} & \palepinkcell{34.3} \\
	& \orangecell{V+T} & \orangecell{\underline{39.9}} & \orangecell{39.0} & \orangecell{41.7} & \orangecell{40.5} & \orangecell{39.5} \\
	\midrule
	\multirow{3}{*}{LLaVA-NeXT-8B~\cite{llavanext} \smaller{(arxiv'24)}} & \lightbluecell{T} & \lightbluecell{23.0} & \lightbluecell{24.1} & \lightbluecell{28.6} & \lightbluecell{27.1} & \lightbluecell{27.6} \\
	& \palepinkcell{V} & \palepinkcell{38.1} & \palepinkcell{36.7} & \palepinkcell{35.0} & \palepinkcell{35.2} & \palepinkcell{35.1} \\
	& \orangecell{V+T} & \orangecell{\underline{39.9}} & \orangecell{\underline{39.2}} & \orangecell{\underline{41.9}} & \orangecell{\underline{40.9}} & \orangecell{\underline{39.7}} \\
	\midrule
	\multirow{3}{*}{InternVL3-8B~\cite{internvl3} \smaller{(arxiv'25)}} & \lightbluecell{T} & \lightbluecell{23.5} & \lightbluecell{25.2} & \lightbluecell{29.6} & \lightbluecell{27.6} & \lightbluecell{28.8} \\
	& \palepinkcell{V} & \palepinkcell{38.5} & \palepinkcell{34.3} & \palepinkcell{35.6} & \palepinkcell{35.6} & \palepinkcell{35.5} \\
	& \orangecell{V+T} & \orangecell{\textbf{40.4}} & \orangecell{\textbf{39.5}} & \orangecell{\textbf{43.5}} & \orangecell{\textbf{42.0}} & \orangecell{\textbf{43.0}} \\
	\bottomrule[1.1pt]
	\end{tabular}
	\label{tab6}
\end{table*}

\begin{figure*}[!t]
	\centering
	\includegraphics[width = \linewidth]{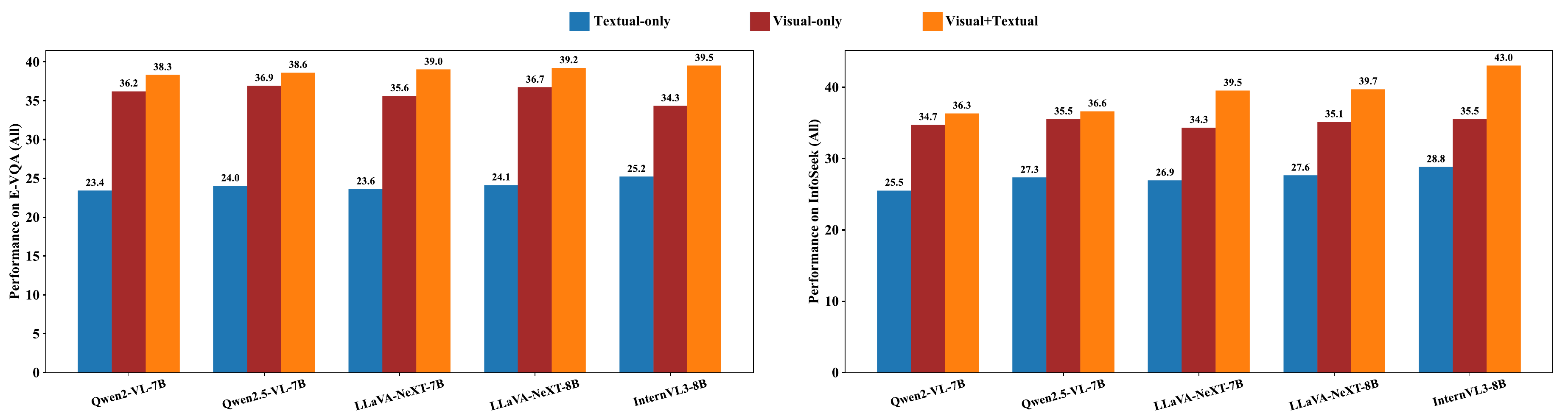}
	\caption{\textbf{Ablation comparison of different retrieval modes across MLLMs.} Results are reported on \textsc{E-VQA}-\emph{All} and \textsc{InfoSeek}-\emph{All}, where ``V'' and ``T'' denote visual and textual features, respectively.}
	\label{fig9}
\end{figure*}
 
\cref{tab4} further compares answer-explanation consistency on \textsc{GQA-REX}. CgRAG achieves the highest Consistency score (97.0) and also leads on Vis. and Tex. scores (3.9 and 4.9), resulting in the best average (4.5). These results suggest that CgRAG improves not only answer accuracy and explanation quality, but also the alignment between predicted answers and their supporting explanations, thereby enhancing reliability in explanatory VQA.

\subsection{Ablation Study}
\label{sec:4.3}

To address \emph{RQ\_3}, we conduct ablation studies to quantify the contributions of the three components in CgRAG and to analyze the impact of key factors.

\subsubsection{Components} 

We first evaluate modular variants to assess the contribution of DCG, EKR, and CPC. As shown in \cref{tab5}, removing all modules yields the lowest performance (23.1 on \textsc{E-VQA}-\emph{Single-Hop} and 24.0 on \textsc{InfoSeek}-\emph{All}), highlighting the necessity of these enhancements. Adding DCG alone substantially improves performance, supporting its role in promoting structured decomposition and CoT-style reasoning. Introducing EKR further boosts accuracy, confirming the importance of CoVQD-guided retrieval for acquiring relevant external evidence. CPC also contributes both independently and synergistically: while EKR+CPC improves upon EKR alone, the full model achieves the best results (40.4 on \textsc{E-VQA}-\emph{Single-Hop} and 43.0 on \textsc{InfoSeek}-\emph{All}). This progressive improvement indicates that each module plays a distinct role, and their combination yields the strongest overall performance.

\subsubsection{Factors} 

We further examine the effects of retrieval mode, the number of QA pairs in CoVQD, and the final prompt structure. As shown in \cref{tab6,fig9}, textual-only (T), visual-only (V), and combined visual-textual (V+T) retrieval are evaluated. V+T consistently performs best across all backbones and datasets, indicating that visual and textual evidence are complementary for knowledge-intensive VQA. Text-only retrieval is markedly weaker, suggesting that textual evidence alone is insufficient for complex visual reasoning, while visual-only retrieval improves performance but remains inferior to V+T fusion.

\begin{table}[!t]
	\centering
	\caption{\textbf{Effect of the number of QA pairs ($K$) in CoVQD.} Bold values indicate the best results.}
	\setlength\tabcolsep{12pt}
	\begin{tabular}{c|cc}
	\toprule[1.1pt]
	{$K$} & \textsc{{E-VQA}} (Single-Hop) & \textsc{{InfoSeek}} (All) \\
	\midrule
	2 & 39.2 & 42.5 \\
	\rowcolor{gray!20}
	4 & \textbf{40.4} & \textbf{43.0} \\
	6 & \textbf{40.4} & 42.9 \\
	\bottomrule[1.1pt]
	\end{tabular}
	\label{tab7}
\end{table}

\begin{figure*}[!t]
	\centering
	\includegraphics[width = 0.9\linewidth]{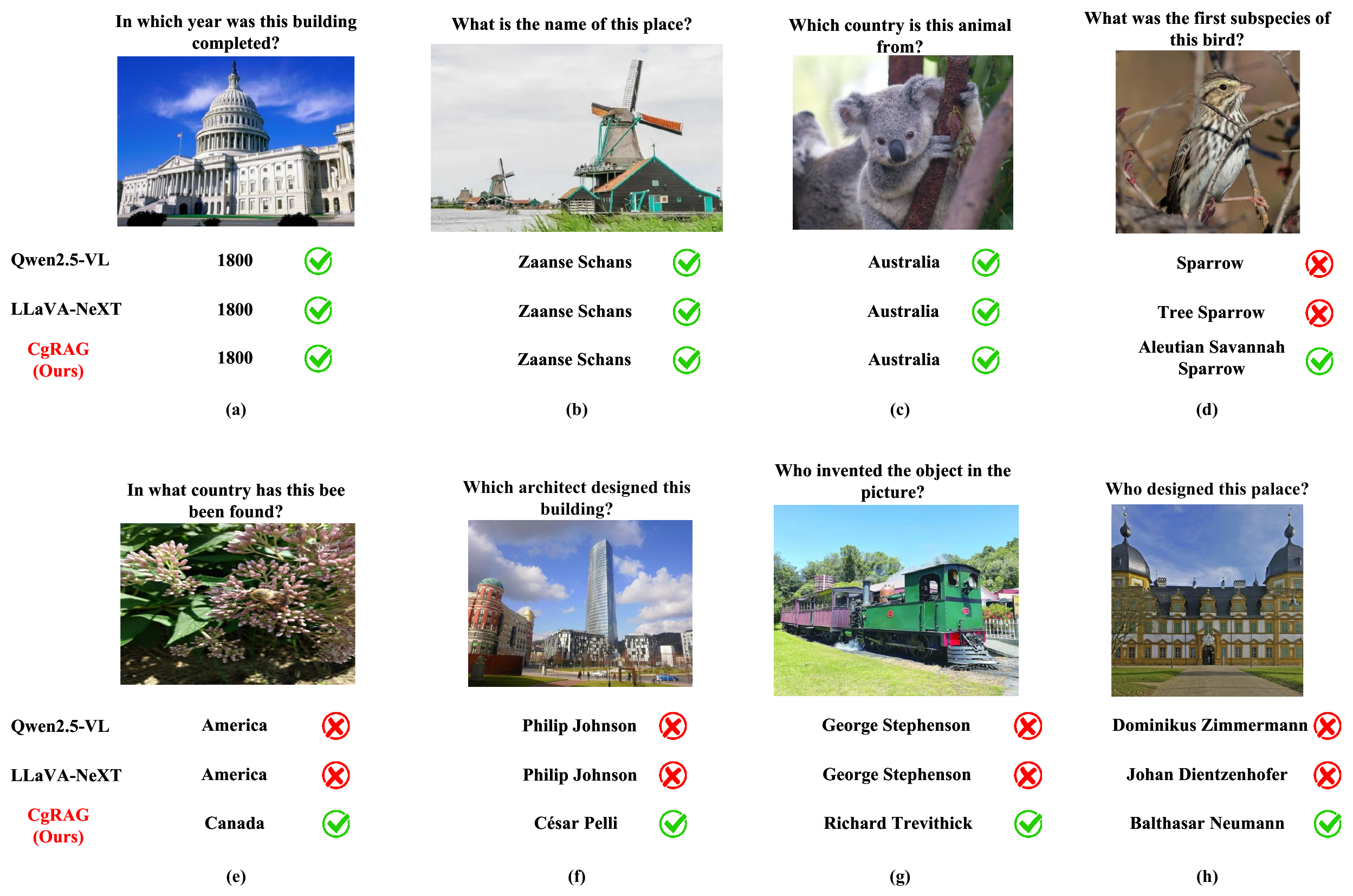}
	\caption{\textbf{Qualitative comparison across different knowledge domains.} Performance differences between Qwen2.5-VL~\cite{qwen2.5vl}, LLaVA-NeXT~\cite{llavanext}, and their CgRAG-enhanced variants are shown on cases from (a-b) commonsense, (c-d) animal, (e) geography, (f) architecture, (g) history, and (h) art domains.}
	\label{fig10}
\end{figure*}

As shown in \cref{tab7}, increasing the number of QA pairs $K$ from 2 to 4 improves performance (39.2 $\to$ 40.4 on \textsc{E-VQA}-\emph{Single-Hop} and 42.5 $\to$ 43.0 on \textsc{InfoSeek}-\emph{All}), indicating that a moderate increase in decomposed context strengthens retrieval alignment and reasoning depth. However, increasing $K$ to 6 does not bring further gains and slightly degrades \textsc{InfoSeek}-\emph{All} (43.0 $\to$ 42.9), suggesting diminishing returns and potential noise from excessive context. These results support $K{=}4$ as a practical trade-off between effectiveness and stability.

\begin{table}[!t]
	\centering
	\scriptsize
	\caption{\textbf{Effect of prompt structure on final performance.} $C$, $I$, and $K$ denote caption, image, and knowledge, respectively. Bold values indicate the best results.}
	\setlength\tabcolsep{12pt}
	\begin{tabular}{c|cc}
	\toprule[1.1pt]
	{Prompt Structure} & \textsc{{E-VQA}} (Single-Hop) & \textsc{{InfoSeek}} (All) \\
	\midrule
	\rowcolor{gray!20}
	C-I-K & \textbf{40.4} & \textbf{43.0} \\
	I-C-K & 40.1 & 39.9 \\
	\bottomrule[1.1pt]
	\end{tabular}
	\label{tab8}
\end{table}

As shown in \cref{tab8}, we compare two prompt structures produced by CPC. The $C$-$I$-$K$ order achieves higher accuracy than $I$-$C$-$K$, indicating that providing a global caption before image patches is more effective for structuring the subsequent integration of retrieved knowledge.

\subsection{Qualitative Analysis}
\label{sec:4.4}

To address \emph{RQ\_4}, we present qualitative comparisons across different domains. \cref{fig10} shows results for Qwen2.5-VL~\cite{qwen2.5vl} and LLaVA-NeXT~\cite{llavanext} with and without CgRAG on eight cases spanning (a-b) commonsense, (c-d) animal, (e) geography, (f) architecture, (g) history, and (h) art. In commonsense cases (a-b) and simple animal case (c), all methods produce the correct answers. However, across the cases (d-h) that require deep domain expertise, CgRAG enables backbone MLLMs to consistently produce correct answers. In contrast, the plain MLLMs are more prone to superficial cues and noisy knowledge integration: in cases (d-e), they rely on shallow textual or visual triggers that lead to incorrect inferences (\emph{e.g.}, ``country'' inducing ``America'', or recognizing a bird and answering ``sparrow''), which hinders evidence-seeking retrieval; in cases (f-h), they produce confused answers due to crude integration of retrieved content (\emph{e.g.}, mismatching ``Philip Johnson'' with the building in the picture despite he was a core figure in the American architectural, or misunderstanding the ``inventor'' as ``reformer'' of the steam locomotive, or confusing the famous work of three Baroque and Rococo artists). These cases illustrate that CgRAG improves fine-grained knowledge integration and supports more reliable inference across diverse domains.

\section{Conclusion}
\label{sec:conclu}

In this work, we propose \underline{\textbf{C}}oVQD-\underline{\textbf{g}}uided \underline{\textbf{RAG}} (CgRAG) to improve MLLM-based VQA on KBVQA tasks. CgRAG unifies Chain-of-Thought (CoT) and Visual Question Decomposition (VQD) into a multi-grained retrieval strategy and introduces Chain-of-VQD (CoVQD) to extract fine-grained multimodal signals from question-image (QI) pairs that guide a precise retrieval-augmented generation process. To effectively integrate the retrieved knowledge within this stepwise guidance, we design a flexible prompting strategy to support MLLM inference. Moreover, to enhance the MLLM’s analytical capability for QI pairs, we develop a fine-tuning paradigm, termed logical implication Direct Preference Optimization (liDPO), within the CgRAG framework. The experiment conducted on various relevant datasets demonstrated the effectiveness of the proposed method.





\balance
\bibliographystyle{ieeetr}

\bibliography{CgRAG}

\end{document}